\documentclass[journal]{IEEEtran}

\usepackage{cite}
\usepackage{amsmath,amssymb,amsfonts}
\usepackage{algorithmic}
\usepackage{graphicx}
\usepackage{textcomp}
\usepackage{xcolor}
\usepackage{hyperref}
\usepackage{url}
\usepackage{booktabs}
\usepackage{multirow}

\hypersetup{
    colorlinks=true,
    linkcolor=blue,
    urlcolor=blue,
    citecolor=blue
}

\renewcommand{\footnoterule}{%
    \kern -3pt
    \hrule width \columnwidth height 0.4pt
    \kern 2pt
}

\def\BibTeX{{\rm B\kern-.05em{\sc i\kern-.025em b}\kern-.08em
    T\kern-.1667em\lower.7ex\hbox{E}\kern-.125emX}}

\begin{document}

\title{HPMD: A Historical Persian Manuscript Dataset for Word Spotting with Line-Level Annotation}

\author{Saeid~Firouzi~Daghigh
        and~Majid~Iranpour Mobarakeh%
\thanks{The authors are with the Department of Computer Engineering
and Information Technology, Payame Noor University, Tehran, Iran
(e-mail: saeedmr881@gmail.com; iranpour@pnu.ac.ir).}}

\markboth{}%
{Firouzi Daghigh and Iranpour Mobarekeh: Historical Persian Handwritten Dataset and Posterior-Based Word Spotting}

\maketitle

\begin{abstract}
Large collections of historical Persian manuscripts have been digitized, but searching them is still slow and mostly manual. Historians usually want to find where a specific name, date, event, or topic appears, which is a word spotting problem rather than a full transcription problem. Progress on this task is limited by two things. First, there is almost no public dataset of historical Persian handwriting; the only notable resource, OpenITI MAKHZAN, contains a relatively small Persian portion. Second, word spotting models usually need word-level bounding boxes, which are very expensive to annotate. In this paper we introduce a new dataset of 223 pages, 3,678 lines, 37,631 words, and 130,630 characters, collected from diverse historical Persian books of poetry and prose and annotated by 11 annotators at the region, line, and text level. We also propose a baseline that is trained only with line-level annotations but returns word-level locations. A fine-tuned line detector finds text lines, and a fine-tuned CRNN recognizer trained with CTC produces a frame-by-character posterior matrix for each line. Instead of decoding the most probable character at each frame, the query is scored directly against this matrix, so visually similar characters in Persian such as \emph{be} and \emph{pe} no longer cause hard failures. The frame alignment also gives the horizontal position of the word inside the line. On the test set, the fine-tuned line detector reaches an F1 of 0.892, and posterior-based search raises the word spotting F1 from 0.487 to 0.558 and recall from 0.340 to 0.548 compared with exact matching on the decoded text, with the decision threshold selected on a held-out validation set. A PHOC attribute-embedding baseline that additionally receives oracle word boundaries at test time reaches an F1 of 0.449, below the proposed method despite its stronger supervision. We also report a distributional analysis of the dataset, a taxonomy of retrieval errors, and a per-condition breakdown of performance. Dataset and codes are available at: \url{https://github.com/saeed5959/persian_handwritten_dataset}
\end{abstract}

\begin{IEEEkeywords}
Historical document analysis, Persian handwriting, word spotting, dataset.
\end{IEEEkeywords}

\section{Introduction}
\IEEEPARstart{P}{ersian} manuscripts form one of the richest written heritages in the world. Libraries and archives have scanned many of them, yet most of these images are not searchable. A historian who wants to know where a particular person, place, date, or event is mentioned still has to read page after page. Automatic tools could shorten this work from months to minutes, but only if they are reliable on the kind of material historians actually use.

A natural first idea is to run an optical character recognition (OCR) or handwritten text recognition (HTR) model and then search the resulting text. This works poorly on historical Persian manuscripts for two reasons. The first is accuracy. Persian script is cursive, letters change shape by position, and many letters share the same base shape and differ only in the number or placement of dots, for example \emph{be}/\emph{pe}/\emph{te}/\emph{se} or \emph{jim}/\emph{che}/\emph{he}/\emph{khe}. In old manuscripts dots are often faded, misplaced, or missing. A standard recognizer outputs only the single most probable character at each position, so one wrong character is enough to make a search fail. The second reason is location. Most OCR pipelines return a string for each line, not the position of each word inside the page, while a historian needs to see exactly where the word is.

Word spotting solves the location problem, but most word spotting methods are trained with word-level bounding boxes \cite{Almazan2014Attributes,Wilkinson2017CtrlF,Krishnan2023HWNetv3,Papazis2025Rerank}. Drawing these boxes is tedious. For a collection of 200 pages one may need to draw around 2,000 line regions but roughly 20,000 word boxes. For low-resource scripts and historical material, this cost is often the real reason there are fewer datasets.

Public data for Persian handwriting is also scarce. Existing Persian datasets such as Sadri \cite{Sadri2016}, Khayyam \cite{Jafarzadeh2024Khayyam}, Hoda \cite{Khosravi2007Hoda}, FHT \cite{Ziaratban2009FHT}, and the recent MPHD \cite{Jampour2026MPHD} were written by modern writers on collection forms. Historical datasets exist for Arabic \cite{Kassis2017VMLHD} and Urdu \cite{Basharat2026UrduKatib}, and OpenITI MAKHZAN \cite{Allen2026Makhzan} covers several Arabic-script languages, but its Persian portion is small.

In this work we address both the data problem and the annotation cost problem. We collected and annotated a dataset of historical Persian manuscripts that covers both poetry and prose, including pages from the \emph{Divan} of Hafez, the \emph{Shahnameh}, and \emph{Majma al-Bayan}. The annotation is done only at the region and line level. We then show that a model trained with this cheaper supervision can still locate individual words. The key idea is to keep the full posterior matrix of a CTC-based recognizer instead of its decoded text, and to search this matrix directly.

The main contributions of this paper are:
\begin{enumerate}
\item A new dataset of historical Persian handwriting with 223 pages, 3,678 lines, 37,631 words, and 130,630 characters, drawn from diverse books of poetry and prose and annotated with text regions, line regions, and transcriptions, together with a distributional analysis of line and word lengths, character frequencies, script styles, genre, illumination, and degradation.
\item A word spotting approach that scores queries against the CTC posterior probability matrix instead of the decoded string. This makes the search tolerant to confusions between similar characters and increases the F1 score over exact matching.
\item A complete and reproducible baseline, including line detection, recognition, an explicit threshold-selection protocol, a comparison against a PHOC attribute-embedding method with stronger supervision, and an error analysis with visual examples.
\end{enumerate}

\section{Related Works}\label{sec:related}

\subsection{Handwriting Datasets}
Public benchmarks have shaped progress in handwriting recognition. For English, the IAM database \cite{Marti2002IAM} remains the most widely used resource for line-level recognition and word spotting. For Arabic-script languages the picture is less complete.

Several Persian handwriting datasets have been released. Hoda \cite{Khosravi2007Hoda} contains a large number of isolated handwritten digits and is limited to digit recognition. The Sadri database \cite{Sadri2016} was collected from 500 writers and includes digits, letters, words, and free text, with writer metadata. Khayyam \cite{Jafarzadeh2024Khayyam} focuses on unconstrained Persian words and contains about 44,000 words, 60,000 letters, and 6,000 digits from 400 writers. FHT \cite{Ziaratban2009FHT} provides line-level Persian text, and MPHD \cite{Jampour2026MPHD} recently added a multi-purpose dataset with 500 writers, line-level transcriptions, isolated characters, and demographic information. All of these datasets were written by modern writers on prepared forms. They do not show the layout, ornaments, script styles, and degradation found in historical manuscripts.

Historical datasets in related scripts are also available. VML-HD \cite{Kassis2017VMLHD} provides 680 pages from five historical Arabic books annotated at the sub-word level and was designed for word spotting and recognition. The Urdu Katib dataset \cite{Basharat2026UrduKatib} contains more than 13,000 text lines written by calligraphers in the Nastaliq style. OpenITI MAKHZAN \cite{Allen2026Makhzan} gathers about 1,500 page images of Arabic, Persian, Ottoman Turkish, and Urdu manuscripts and prints with line-level segmentation and transcription. It is the closest resource to ours, but only a limited part of it is Persian, and it is mainly intended for training transcription models. Table~\ref{tab:datasets} compares these datasets with ours.

\subsection{Handwritten Text Recognition}
Modern HTR systems usually combine a convolutional feature extractor with a recurrent sequence model and a Connectionist Temporal Classification (CTC) output layer. Wang and Hu \cite{Wang2017GRCNN} showed that adding gated recurrent connections inside the convolutional layers improves context modeling for text recognition. Transformer-based models are now common as well. HATFormer \cite{Chan2024HATFormer} adapts a transformer encoder-decoder to historical handwritten Arabic, and a recent study on TrOCR for medieval manuscripts \cite{Sharma2026TrOCR} shows that fine-tuning choices strongly affect accuracy on small historical datasets. These systems output a text string, and their accuracy is usually reported with character or word error rates. For search, however, a single wrong character in the output is enough to miss a word.

\subsection{Layout Analysis}
Line detection is a key step in historical document processing. Kiessling \cite{Kiessling2020BLLA} proposed a modular system that predicts baselines and regions with a neural network and then computes line polygons. This system, available in the kraken engine, handles curved and rotated lines and has been evaluated on Arabic-script manuscripts. We use it as the starting point for our line detector.

\subsection{Word Spotting}
Word spotting retrieves the occurrences of a query in document images without full transcription. The attribute-embedding framework of Almaz\'{a}n \emph{et al.} \cite{Almazan2014Attributes} introduced the Pyramidal Histogram of Characters (PHOC), a fixed-length binary code that records which characters appear in which relative part of a word. Because a PHOC can be computed both from a word image and from a query string, images and strings live in a common space and retrieval reduces to a nearest-neighbour search. PHOCNet \cite{Sudholt2016PHOCNet} replaced the hand-crafted features with a convolutional network and became the standard query-by-string baseline. Ctrl-F-Net \cite{Wilkinson2017CtrlF} extended this to segmentation-free spotting on full pages by jointly proposing word regions and embedding them. HWNet~v3 \cite{Krishnan2023HWNetv3} learns a joint embedding of word images and text supporting both retrieval and lexicon-based recognition, and Papazis \emph{et al.} \cite{Papazis2025Rerank} re-rank retrieved word images using semantic embeddings from language models. These methods give strong results, but they all rely on word-level boxes for training or evaluation. Our approach needs only line-level annotation and recovers the word position from the frame alignment of the recognizer. In Section~\ref{sec:qbs} we compare against a PHOC baseline that is given oracle word boundaries at test time, as an upper bound for the stronger-supervision family.

\begin{table}[!t]
\centering
\caption{Comparison of our dataset with related handwriting datasets.}
\label{tab:datasets}
\begin{tabular}{lccc}
\toprule
Dataset & Persian & Handwritten & Historical \\
\midrule
VML-HD \cite{Kassis2017VMLHD} & No & Yes & Yes \\
Urdu Katib \cite{Basharat2026UrduKatib} & No & Yes & Yes \\
IAM \cite{Marti2002IAM} & No & Yes & No \\
Sadri \cite{Sadri2016} & Yes & Yes & No \\
Khayyam \cite{Jafarzadeh2024Khayyam} & Yes & Yes & No \\
Hoda \cite{Khosravi2007Hoda} & Yes & Yes & No \\
FHT \cite{Ziaratban2009FHT} & Yes & Yes & No \\
MPHD \cite{Jampour2026MPHD} & Yes & Yes & No \\
OpenITI MAKHZAN \cite{Allen2026Makhzan} & Yes & Yes & Yes \\
Ours & Yes & Yes & Yes \\
\bottomrule
\end{tabular}
\end{table}

\section{Dataset}\label{sec:dataset}

\subsection{Source Material}
The pages were selected carefully from historical Persian books so that the dataset covers different genres, scripts, and layouts. The sources include poetry and prose. Poetry pages often use a two-column layout with separated hemistichs, while prose pages have long, continuous lines. The pages also vary in writing style, ink color (for example, red headings), decoration, and state of preservation. Some pages contain illuminated frames, cloud-shaped gilded backgrounds around the text, stains, and faded strokes. All pages were scanned at high quality and stored in JPG format.

\subsection{Annotation Protocol}
\begin{figure*}[!t]
\centering
\includegraphics[width=0.85\textwidth]{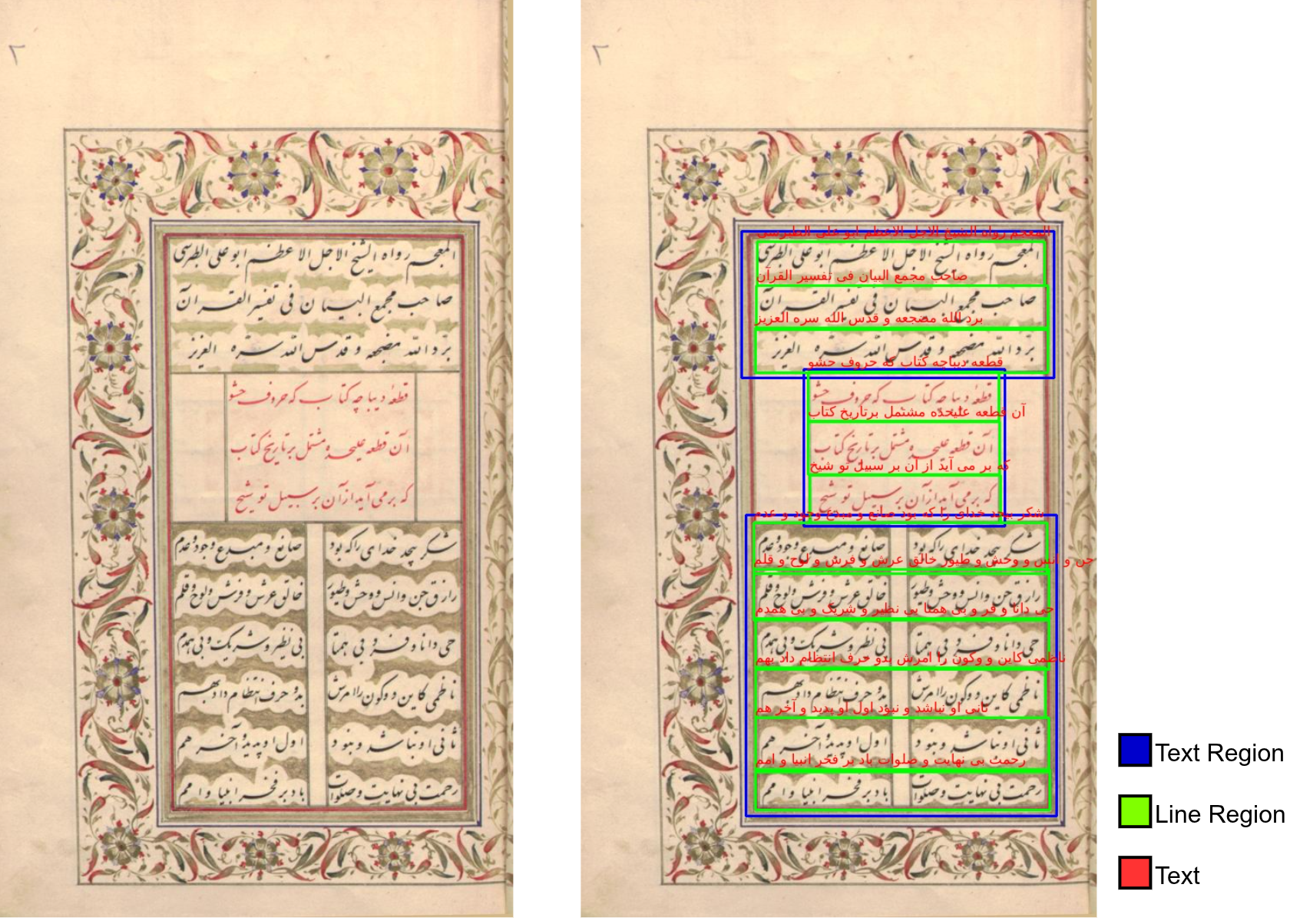}
\caption{An example page from the dataset (\emph{Majma al-Bayan}). Left: the original scanned page. Right: the annotation made in Transkribus, with text regions in blue, line regions in green, and the line transcriptions in red. The page contains a decorated frame, red rubrics, and a two-column poetry block, which illustrate the layout variety of the dataset.}
\label{fig:annotation}
\end{figure*}

Annotation was carried out by 11 annotators using the Transkribus platform\footnote{\url{https://www.transkribus.org/}}. For each page the annotators marked three levels of information: text regions, line regions within each text region, and the transcription of each line. Fig.~\ref{fig:annotation} shows an example. Each transcription was entered carefully to match the written text, and the annotations were exported as XML files. We did not draw word-level boxes. This choice kept the annotation cost manageable, and, as shown in Section~\ref{sec:method}, word positions can still be recovered by the model.

In addition to the per-line annotation, each page carries five page-level attributes that were assigned by visual inspection: genre (poetry, prose, mixed), script style (\emph{shekasteh}, \emph{nastaliq}, \emph{naskh}), degradation (none, light, moderate), illumination or \emph{tazhib} (none, border, both border and gilded cloud background), and a subjective reading difficulty (easy, moderate, hard) reflecting how hard a trained reader finds the hand to decipher. These attributes are released with the dataset and are used in Section~\ref{sec:difficulty} to report performance by condition.

\subsection{Statistics and Splits}
The dataset contains 223 pages, 3,678 lines, 37,631 words, and 130,630 characters (Table~\ref{tab:stats}). After the normalization described below, 3,656 lines carry a non-empty transcription; the remaining lines are annotated regions whose content reduces to punctuation or decoration only, and they are excluded from training and evaluation. On average a page has about 16.4 lines and a line has about 10.3 words. We split the data at the page level into training, validation, and test sets, so no page appears in more than one set. The split is given in Table~\ref{tab:split}.

\begin{table}[!t]
\centering
\caption{Overall statistics of the dataset.}
\label{tab:stats}
\begin{tabular}{cccc}
\toprule
Pages & Lines & Words & Characters \\
\midrule
223 & 3,678 & 37,631 & 130,630 \\
\bottomrule
\end{tabular}
\end{table}

\begin{table}[!t]
\centering
\caption{Page-level split of the dataset used for training and evaluation.}
\label{tab:split}
\begin{tabular}{lccc}
\toprule
Split & Pages & Lines & Words \\
\midrule
Train & 144 & 2,329 & 24,301 \\
Validation & 35 & 614 & 6,079 \\
Test & 44 & 735 & 7,251 \\
\midrule
Total & 223 & 3,678 & 37,631 \\
\bottomrule
\end{tabular}
\end{table}

\subsection{Text Normalization}
Some symbols appear rarely in the transcriptions and make training harder without helping search. Before training, we normalize the text by removing diacritics (short vowel marks and \emph{tashdid}), mapping the different forms of \emph{alef} to a single form, mapping the Arabic \emph{yeh} and \emph{kaf} to their Persian counterparts, replacing the zero-width non-joiner with a space, and removing punctuation and other rare symbols. The same normalization is applied to the search queries, so a user can type a word in its plain form and still match the manuscript.

\subsection{Distributional Analysis}\label{sec:distribution}
\begin{figure*}[!t]
\centering
\includegraphics[width=0.95\textwidth]{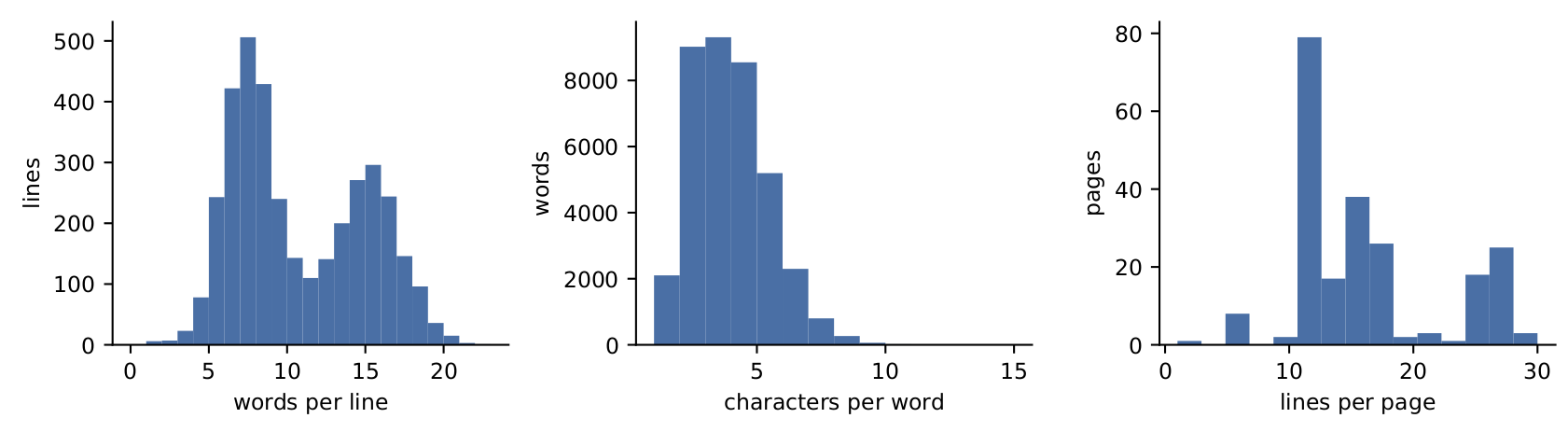}
\caption{Distribution of words per line (left), characters per word (middle), and lines per page (right). The bimodal shape of the first and third histograms comes from the mixture of two-column poetry pages and full-width prose pages.}
\label{fig:lengths}
\end{figure*}

A dataset description is only useful for comparison if the distribution of its content is known. Table~\ref{tab:dist} summarizes the structural quantities and Fig.~\ref{fig:lengths} shows their histograms.

The distribution of words per line is clearly bimodal, with one mode near 7-8 words and a second near 15-16. This directly reflects the genre mixture: poetry pages use a two-column layout in which each annotated line is a single hemistich, while prose pages run the full width of the text block. Any method evaluated on this dataset therefore has to cope with a factor-of-two variation in line length, and results averaged over all lines mix two rather different regimes. Words are short, with a median of 3 characters and a mean of 3.5, which is important for retrieval: as shown in Section~\ref{sec:error}, short queries are the dominant source of false positives.

\begin{table}[!t]
\centering
\caption{Distribution of structural quantities in the dataset.}
\label{tab:dist}
\begin{tabular}{lccccc}
\toprule
Quantity & Mean & Std & Median & Min & Max \\
\midrule
Lines per page & 16.4 & 6.3 & 15 & 1 & 30 \\
Words per line & 10.3 & 4.2 & 9 & 1 & 22 \\
Characters per line & 35.7 & 14.6 & 28 & 3 & 78 \\
Characters per word & 3.5 & 1.5 & 3 & 1 & 15 \\
\bottomrule
\end{tabular}
\end{table}

\begin{figure*}[!t]
\centering
\includegraphics[width=0.95\textwidth]{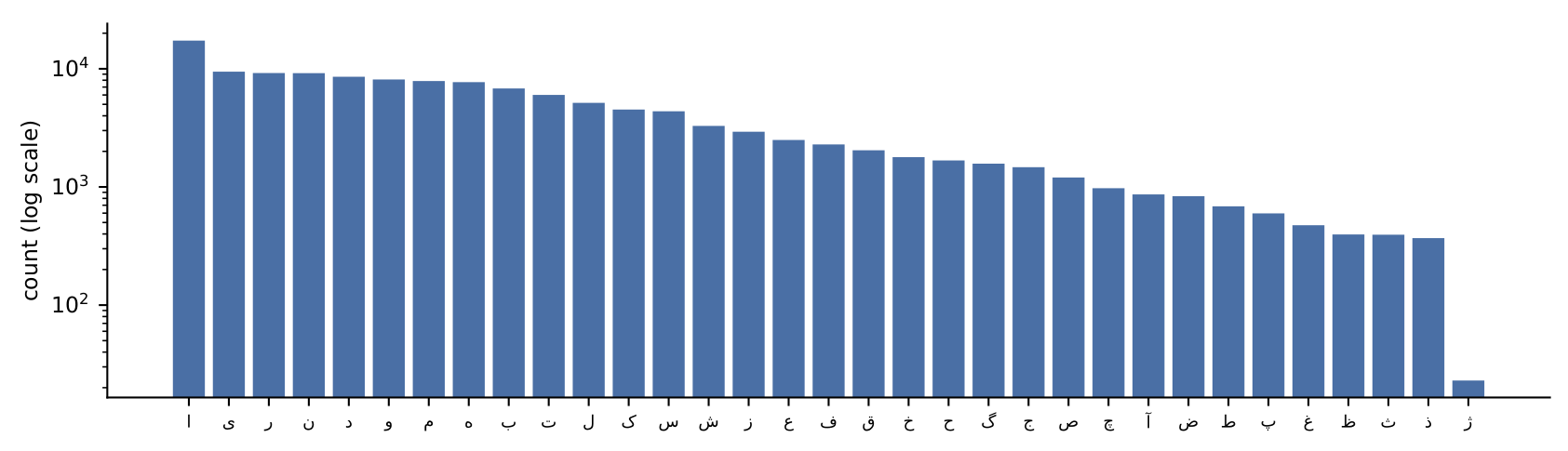}
\caption{Character frequency in the dataset after normalization, on a logarithmic scale. The distribution spans almost three orders of magnitude, from \emph{alef} with about 17,000 occurrences to \emph{zhe} with 23.}
\label{fig:charfreq}
\end{figure*}

The character frequency distribution (Fig.~\ref{fig:charfreq}, Table~\ref{tab:chars}) spans nearly three orders of magnitude. The most frequent letter, \emph{alef}, occurs about 17,000 times, while \emph{zhe} occurs only 23 times in the whole dataset. Seven letters account for less than 0.5\% of all characters each. This long tail is a practical constraint on any recognizer trained on this data: a character seen 23 times cannot be learned reliably, and queries containing it should be expected to fail. We report this explicitly so that future work can separate genuine modelling improvements from differences in how rare characters are handled.

\begin{table}[!t]
\centering
\caption{Frequency of the least frequent Persian letters after normalization. Counts are over 130,628 characters.}
\label{tab:chars}
\begin{tabular}{lccc}
\toprule
Letter & Unicode & Count & Share (\%) \\
\midrule
zhe & U+0698 & 23 & 0.018 \\
zal & U+0630 & 369 & 0.282 \\
se & U+062B & 394 & 0.302 \\
za & U+0638 & 396 & 0.303 \\
ghayn & U+063A & 474 & 0.363 \\
pe & U+067E & 597 & 0.457 \\
zad & U+0636 & 835 & 0.639 \\
che & U+0686 & 975 & 0.746 \\
sad & U+0635 & 1,202 & 0.920 \\
gaf & U+06AF & 1,573 & 1.204 \\
\bottomrule
\end{tabular}
\end{table}

\begin{figure*}[!t]
\centering
\includegraphics[width=0.95\textwidth]{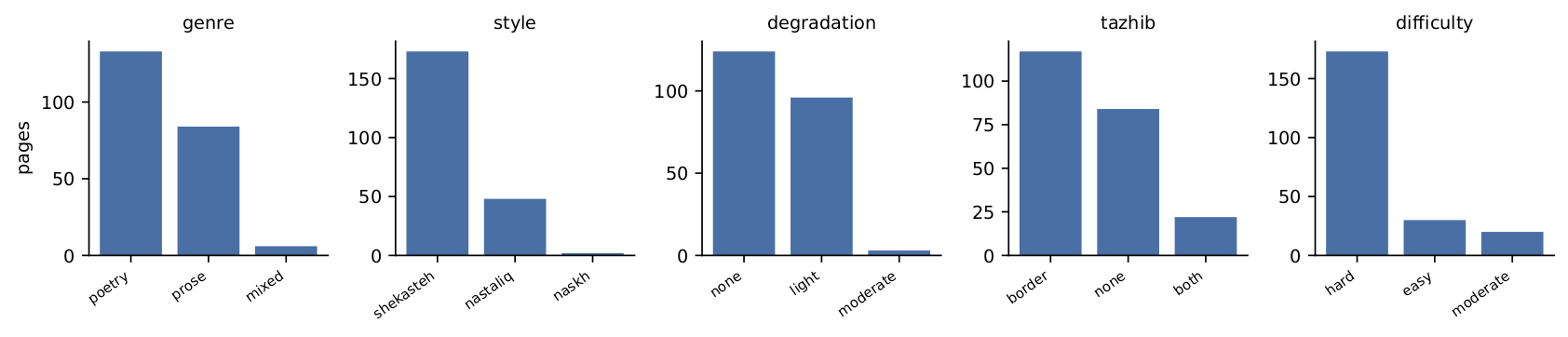}
\caption{Composition of the dataset by genre, script style, degradation, illumination (\emph{tazhib}), and subjective reading difficulty, in pages. The difficulty label coincides with the \emph{shekasteh} script class, since that script drives the reading difficulty on almost every page where it appears.}
\label{fig:composition}
\end{figure*}

Table~\ref{tab:meta} and Fig.~\ref{fig:composition} give the composition by page attribute. Poetry accounts for 133 of the 223 pages and prose for 84. The dominant script is \emph{shekasteh} (173 pages), followed by \emph{nastaliq} (48) and \emph{naskh} (2). \emph{Shekasteh} is the hardest of the three to read, since letters are heavily joined and many dots are omitted, and its dominance is a deliberate choice: it is the script in which a large part of the surviving Persian administrative and literary manuscript record is written, and it is under-represented in existing datasets. One hundred and thirty-nine pages carry some illumination, and 99 show some degradation. Two categories are very small, \emph{naskh} with 2 pages and moderate degradation with 3, and results conditioned on them should not be treated as reliable.

\begin{table}[!t]
\centering
\caption{Composition of the dataset by genre, script style, degradation, illumination, and subjective reading difficulty.}
\label{tab:meta}
\begin{tabular}{llccc}
\toprule
Attribute & Value & Pages & Lines & Words \\
\midrule
Genre & poetry & 133 & 2,535 & 22,032 \\
 & prose & 84 & 1,033 & 14,660 \\
 & mixed & 6 & 88 & 939 \\
\midrule
Style & shekasteh & 173 & 3,000 & 32,878 \\
 & nastaliq & 48 & 633 & 4,443 \\
 & naskh & 2 & 23 & 310 \\
\midrule
Degradation & none & 124 & 2,293 & 19,639 \\
 & light & 96 & 1,328 & 17,545 \\
 & moderate & 3 & 35 & 447 \\
\midrule
Tazhib & none & 84 & 973 & 13,944 \\
 & border & 117 & 2,310 & 21,464 \\
 & both & 22 & 373 & 2,223 \\
\midrule
Difficulty & easy & 30 & 301 & 2,690 \\
 & moderate & 20 & 355 & 2,063 \\
 & hard & 173 & 3,000 & 32,878 \\
\bottomrule
\end{tabular}
\end{table}

\section{Methodology}\label{sec:method}

\begin{figure*}[!t]
\centering
\includegraphics[width=0.85\textwidth]{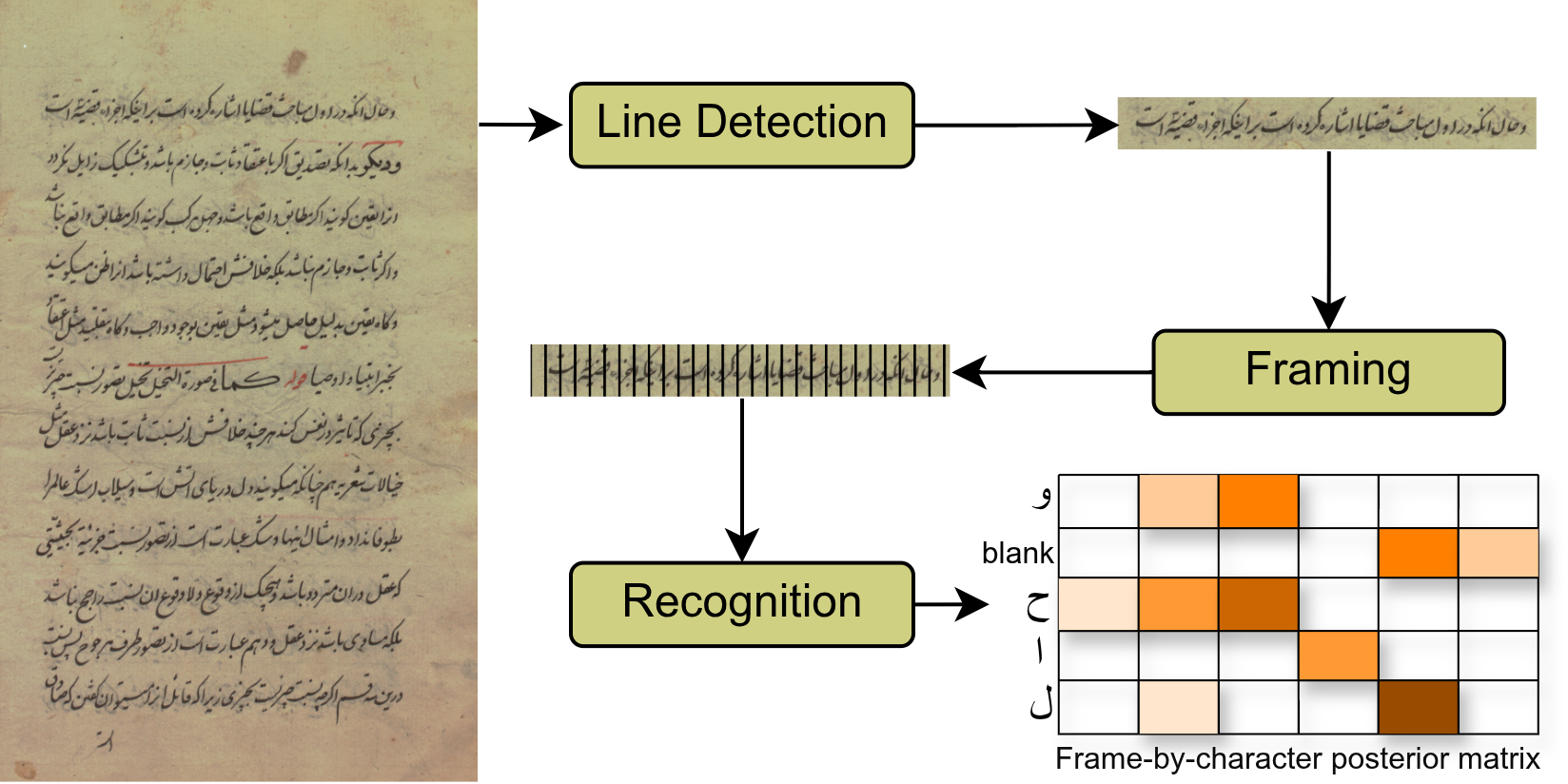}
\caption{Overview of the proposed pipeline. Text lines are detected on the page, each line image is split into frames by the recognizer, and the recognizer outputs a frame-by-character posterior matrix (including the CTC blank). Word spotting is performed on this matrix instead of on the decoded text, and the matched frames give the position of the word in the line and the page.}
\label{fig:pipeline}
\end{figure*}

The proposed pipeline has three stages: line detection, recognition, and posterior-based search (Fig.~\ref{fig:pipeline}). All models were trained with the kraken engine\footnote{\url{https://kraken.re}}.

\subsection{Line Detection}
For line detection we fine-tune the baseline and layout analysis model (BLLA) of Kiessling \cite{Kiessling2020BLLA}. The network predicts baseline and region heatmaps for the page, and the baselines are then turned into line polygons. We start from the default pretrained BLLA weights and fine-tune them on the line annotations of our training set. Each detected line is cropped and rectified into a straight line image for the recognizer.

\subsection{Recognition}
For recognition we fine-tune the pretrained \texttt{all\_arabic} model\footnote{\url{https://zenodo.org/records/7050270}}, a CRNN trained with CTC loss on Arabic-script text. The model follows the usual convolutional-recurrent design \cite{Wang2017GRCNN}. A line image is rescaled to a fixed height while its width is kept free. Convolutional layers downsample the image along its width and produce a sequence of $T$ frames, where each frame describes a thin vertical slice of the line. A bidirectional LSTM reads this sequence in both directions, and a linear layer with softmax gives, for every frame, a probability distribution over the character set plus a special blank symbol. CTC training needs only the line transcription as a label, so no character or word boxes are required. Because Persian is written from right to left while frames are ordered from left to right in the image, kraken applies the Unicode bidirectional algorithm to put the labels in display order. We apply the same reordering to the queries.

Both models were trained with a learning rate of $10^{-4}$, a batch size of 8, and 40 epochs.

\subsection{The Posterior Matrix}
For a line image with $T$ frames and a character set with $K$ characters, the recognizer outputs a matrix
\begin{equation}
Y \in [0,1]^{T \times (K+1)}, \qquad \sum_{c} y_t(c) = 1,
\end{equation}
where $y_t(c)$ is the probability of symbol $c$ at frame $t$ and the extra column is the blank. A normal OCR system reduces $Y$ to text by picking $\arg\max_c y_t(c)$ at every frame and then removing repeated symbols and blanks. We keep $Y$ in full. In practice we read the values of the layer before the final decoding step, i.e., the softmax output of the network, and store one matrix per detected line.

\subsection{Scoring a Query Against the Posterior}
Let the normalized query be $q = (q_1, \dots, q_L)$. As in CTC, we insert blanks around and between its characters to get the extended sequence
\begin{equation}
q' = (\varnothing, q_1, \varnothing, q_2, \dots, \varnothing, q_L, \varnothing),
\end{equation}
of length $2L+1$, where $\varnothing$ is the blank. The query can appear anywhere in the line, so the alignment is allowed to start and end at any frame.

To make scores comparable across lines, we measure each frame relative to the best symbol at that frame:
\begin{equation}
r_t(c) = \log y_t(c) - \max_{k} \log y_t(k) \;\le\; 0 .
\end{equation}
If $c$ is the top symbol at frame $t$, then $r_t(c)=0$; otherwise it is negative, and the more the model prefers another symbol, the more negative it becomes.

We then find the best CTC alignment of $q'$ with a Viterbi recursion. Let $\delta_t(s)$ be the best score of a path that ends at frame $t$ in state $s$ of $q'$:
\begin{equation}
\delta_t(s) = r_t(q'_s) + \max\bigl\{\delta_{t-1}(s),\; \delta_{t-1}(s-1),\; \delta_{t-1}(s-2)\bigr\},
\end{equation}
where the jump from $s-2$ is allowed only when $q'_s$ is not blank and $q'_s \neq q'_{s-2}$, as in standard CTC. To let the word start at any frame, the first two states ($s=1$ and $s=2$) may also start fresh with a previous score of zero. The final score of the query in the line is
\begin{equation}
S(q) = \max_{t} \max\bigl\{\delta_t(2L),\, \delta_t(2L+1)\bigr\},
\end{equation}
and we normalize it to a per-character log score,
\begin{equation}
s(q) = \frac{S(q)}{L} \;\le\; 0 ,
\end{equation}
equivalently a per-character confidence $C(q) = \exp(s(q)) \in (0,1]$. Dividing by $L$ keeps long queries from being penalized only for their length. A line is reported as containing the query when $s(q) \ge \tau$. The selection of $\tau$ on the validation set is described in Section~\ref{sec:threshold}. Several non-overlapping matches can be returned from the same line.

\subsection{Why Posterior Scoring Helps: An Intuitive View}
It is easiest to see the benefit with an example. Suppose the manuscript contains the word \emph{ketab} (``book''), which ends with the letter \emph{be}. Because the dot is faint, the recognizer is not sure about the last letter. At that frame it gives \emph{be} a probability of 0.45 and \emph{pe} a probability of 0.40.

A normal OCR system picks the highest value at each frame and writes down the result. If the dot is read slightly differently and \emph{pe} wins, the output becomes a different string, and a text search for \emph{ketab} finds nothing. The word is lost even though the model was almost right.

Posterior scoring does not force this early choice. It asks a different question: ``how likely is it that the word \emph{ketab} is written here?'' Since \emph{be} still has a high probability at that frame, the query still receives a high score and the line is returned. In other words, the uncertainty of the model is kept until the end, and small mistakes reduce the score a little instead of removing the word completely. This matters most for Persian, where many letters differ only by dots.

The same idea explains why recall increases. A historian mainly needs the correct occurrences to appear among the results; the recognizer does not have to be perfect at every character. The price is that some visually similar words also receive a reasonable score, which can lower precision. We quantify this trade-off in Section~\ref{sec:error}.

\subsection{Recovering the Word Position}
The alignment also tells us where the word is. The best path gives the first frame $t_s$ and the last frame $t_e$ that are assigned to the query characters. Since each frame covers a fixed slice of the line image, the horizontal extent of the word in a line image of width $W$ is approximately
\begin{equation}
x_{\text{start}} = \frac{t_s}{T}\,W, \qquad x_{\text{end}} = \frac{t_e+1}{T}\,W .
\end{equation}
The vertical extent is taken from the detected line. By mapping these coordinates back through the line polygon, we obtain the word location on the page. In this way a model trained only with line annotations produces word-level results. Fig.~\ref{fig:qualitative}(c)-(d) shows two examples of the recovered box.

\subsection{Exact-Match Baseline}
To measure the effect of the posterior search, we compare it with a standard decode-then-search baseline. Here the line is decoded with the argmax path, and a hit is reported when the query appears as a word in the decoded text. The position is taken from the frames of the decoded characters, in the same way as above.

\subsection{PHOC Query-by-String Baseline}\label{sec:phoc}
As a representative of the stronger-supervision family we implement a PHOC attribute-embedding baseline in the style of Almaz\'{a}n \emph{et al.} \cite{Almazan2014Attributes} and PHOCNet \cite{Sudholt2016PHOCNet}. A word is encoded as a binary pyramidal histogram over five levels ($1, 2, 3, 4, 5$ splits) of a 42-symbol alphabet, giving a 630-dimensional target. A convolutional network with four blocks of two $3\times3$ convolutions followed by max pooling, and two fully connected layers, maps a word image to this space and is trained with a binary cross-entropy loss. At query time the PHOC of the query string is computed analytically and compared to every word embedding by cosine similarity; a line is scored by the maximum similarity over its words.

This baseline needs word images, which our dataset does not provide. We obtain them by CTC forced alignment of the \emph{ground-truth} transcription against the posterior matrix, which yields a frame span per character and therefore a box per word. This is done on both the training and the test pages. Consequently the PHOC baseline receives information that the proposed method never uses, namely the correct transcription of every test line, and its result should be read as an upper bound for word-box-supervised methods on this data rather than as a like-for-like comparison.

\section{Experiments and Results}\label{sec:results}

\subsection{Evaluation Protocol}
\emph{Line detection.} A predicted line is matched to a ground-truth line when their intersection over union (IoU) is at least 0.3, with each ground-truth line matched at most once. We report precision, recall, F1, and the mean IoU of the matched lines on the 44 test pages (735 lines).

\emph{Word spotting.} For each test page, ten words of at least three characters were chosen at random from the ground truth, giving 396 distinct query words. Every query is searched in all lines of all test pages, producing roughly $2.8\times10^{5}$ query-line pairs of which 1,430 are relevant. A retrieved result is counted as correct when it falls in a line that contains the query word in its ground-truth transcription. We report precision, recall, and F1 over all queries at the operating threshold $\tau$.

We do not use character or word error rate here. Our goal is to find words, not to produce a perfect transcription, and a model with a moderate error rate can still be useful for retrieval if the correct words receive high scores.

\emph{Models.} ``Base model'' refers to the pretrained BLLA and \texttt{all\_arabic} models without fine-tuning. ``Our model'' refers to the same models after fine-tuning on our training set.

\subsection{Line Detection Results}
Table~\ref{tab:line} shows the line detection results. Fine-tuning improves all measures. Precision increases from 0.751 to 0.867 and recall from 0.841 to 0.920, and the F1 score rises from 0.793 to 0.892. The mean IoU also increases from 0.412 to 0.534. The mean IoU values are moderate because the line polygons produced from baselines do not have exactly the same extent as the manually drawn line regions, especially for lines with tall ascenders, long descenders, or decorated backgrounds. A threshold of 0.3 is still enough to decide whether the correct line was found, which is what matters for retrieval.

\begin{table}[!t]
\centering
\caption{Line detection results on the test set (44 pages, 735 lines) at an IoU threshold of 0.3.}
\label{tab:line}
\begin{tabular}{lcccc}
\toprule
Model & Precision & Recall & F1 & Mean IoU \\
\midrule
Base model & 0.751 & 0.841 & 0.793 & 0.412 \\
Our model & \textbf{0.867} & \textbf{0.920} & \textbf{0.892} & \textbf{0.534} \\
\bottomrule
\end{tabular}
\end{table}

\subsection{Threshold Selection}\label{sec:threshold}
The decision threshold $\tau$ is selected by maximizing F1 on the 35 validation pages and is then frozen before the test set is touched. This gives $\tau^{*} = -0.80$ in per-character log score, equivalently a per-character confidence of $0.45$. At this threshold the method reaches F1 $=0.572$ on validation and F1 $=0.558$ on test. The best threshold that could have been chosen with knowledge of the test set is $\tau=-0.88$, giving F1 $=0.565$. The gap of $0.007$ F1 between the validation-selected and the oracle threshold indicates that $\tau$ transfers well across pages, which matters in practice because a user of the released system cannot tune it.

\begin{figure*}[!t]
\centering
\includegraphics[width=0.95\textwidth]{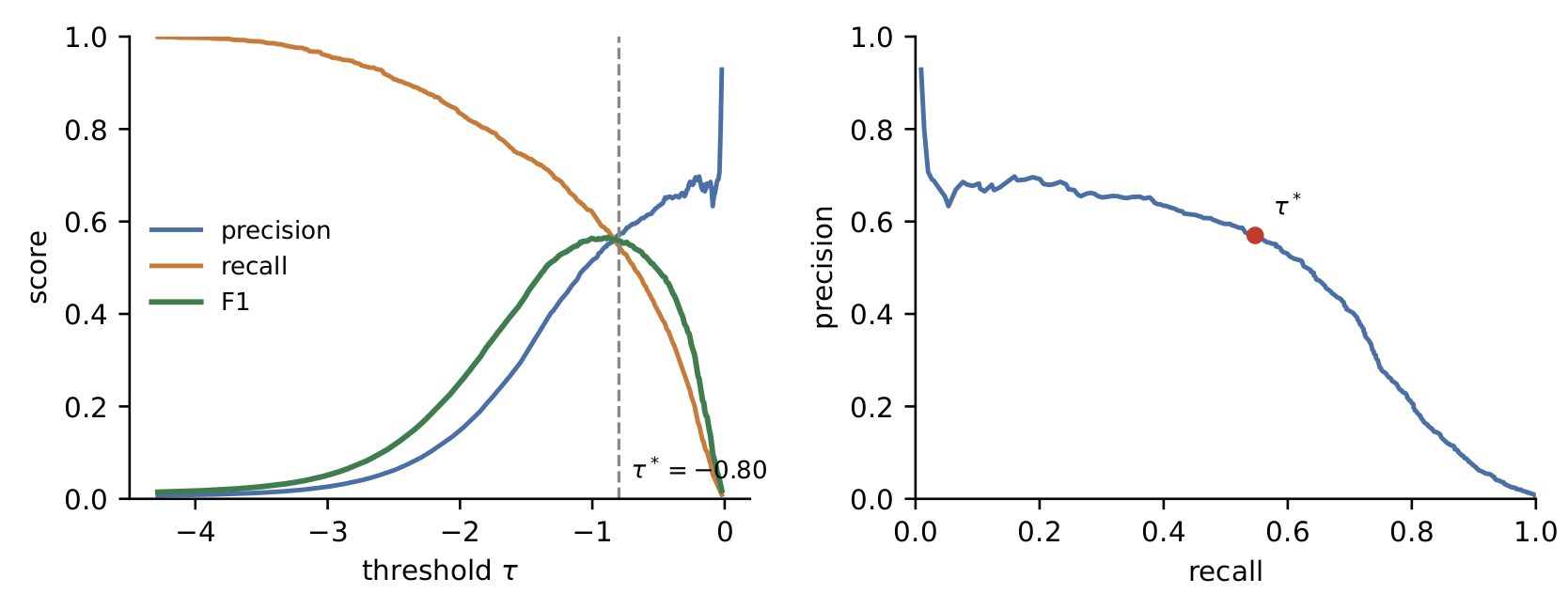}
\caption{Effect of the decision threshold $\tau$ on the test set. Left: precision, recall, and F1 as functions of $\tau$; the dashed line marks $\tau^{*}=-0.80$, selected by maximizing F1 on the validation set. Right: the corresponding precision-recall curve, with $\tau^{*}$ marked. Precision only rises sharply in the last fraction of the range, where recall has already collapsed.}
\label{fig:threshold}
\end{figure*}

Fig.~\ref{fig:threshold} shows precision, recall, and F1 as functions of $\tau$ on the test set, together with the precision-recall curve. The F1 curve is flat within about $\pm0.3$ of $\tau^{*}$, so the exact value is not critical. The two ends of the curve are more informative than the maximum. Precision rises steeply only in the last fraction of the range: reaching precision $0.90$ requires $\tau=-0.02$, which retrieves almost nothing (recall $0.009$). Conversely, recall $0.90$ is available at $\tau=-2.42$ but at precision $0.071$, meaning roughly thirteen results must be inspected per correct hit. Table~\ref{tab:threshold} lists the three operating points.

This shape has a direct consequence for how the system should be deployed. The near-vertical precision curve at the right end means there is no threshold at which the method is both precise and useful; a historian must either accept a moderate precision around $0.57$ at the F1 optimum, or work in a high-recall regime and filter visually. The latter is often acceptable, because judging a returned line image takes a second, whereas a missed occurrence is invisible.

\begin{table}[!t]
\centering
\caption{Operating points of posterior search on the test set. $\tau^{*}$ is selected by maximizing F1 on the validation set and is not tuned on test.}
\label{tab:threshold}
\begin{tabular}{lcccc}
\toprule
Operating point & $\tau$ & Precision & Recall & F1 \\
\midrule
Precision $\geq$ 0.90 & $-0.02$ & 0.929 & 0.009 & 0.018 \\
$\tau^{*}$ (max F1 on val) & $-0.80$ & 0.570 & 0.548 & \textbf{0.558} \\
Recall $\geq$ 0.90 & $-2.42$ & 0.071 & 0.900 & 0.132 \\
\bottomrule
\end{tabular}
\end{table}

\subsection{Word Spotting Results}
Table~\ref{tab:spotting} reports the word spotting results for the two models and the two search strategies.

With the base model and exact matching, recall is only 0.043. The pretrained Arabic recognizer rarely produces Persian words exactly, so the search almost never succeeds, even though the few hits it finds are often correct (precision 0.627). Scoring against the posterior raises recall to 0.218 and the F1 score from 0.080 to 0.248, about three times higher. This shows that useful information is present in the posterior even when the decoded text is wrong.

Fine-tuning improves both strategies. With exact matching, our model reaches a precision of 0.857 but a recall of only 0.340: when the decoded word is correct it is almost always a real occurrence, but about two thirds of the occurrences are missed because at least one character was decoded incorrectly. With posterior search at the validation-selected threshold, recall increases to 0.548 and F1 from 0.487 to 0.558, while precision drops to 0.570. This is the expected trade-off, and Section~\ref{sec:error} identifies where the lost precision goes.

\begin{table}[!t]
\centering
\caption{Word spotting results on the test set (44 pages, 735 lines, 7,251 words) with 396 query words. Posterior search uses $\tau^{*}=-0.80$ selected on the validation set.}
\label{tab:spotting}
\begin{tabular}{llccc}
\toprule
Model & Search & Precision & Recall & F1 \\
\midrule
\multirow{2}{*}{Base model} & Exact OCR & 0.627 & 0.043 & 0.080 \\
 & Posterior & 0.288 & 0.218 & 0.248 \\
\midrule
\multirow{2}{*}{Our model} & Exact OCR & \textbf{0.857} & 0.340 & 0.487 \\
 & Posterior & 0.570 & \textbf{0.548} & \textbf{0.558} \\
\bottomrule
\end{tabular}
\end{table}

\subsection{Comparison with a Query-by-String Method}\label{sec:qbs}
Table~\ref{tab:qbs} compares the proposed method with the PHOC baseline of Section~\ref{sec:phoc}. The PHOC network reaches F1 $=0.449$ and mAP $=0.266$ at line level, and F1 $=0.462$ and mAP $=0.393$ at page level. Its precision at the F1 optimum is higher than ours (0.664 against 0.570), which is expected: it scores whole words against whole words and therefore does not fire on substrings. Its recall is much lower (0.339 against 0.548).

The comparison should be read with two qualifications, both of which favour the baseline. First, it is given oracle word boundaries derived from the ground-truth transcription of the test lines, so its segmentation is perfect in a way no deployed system could be. Second, our implementation is a faithful but not exhaustively tuned PHOCNet; the original works train for far longer with heavier augmentation, and a fully tuned version would score higher. Even with these advantages the attribute embedding does not overtake posterior search at this data scale. We read this as a statement about data rather than about architecture: learning a 630-dimensional image-to-attribute mapping from roughly 19,000 word crops of a single script family is a harder estimation problem than adapting an already-pretrained CTC recognizer, whose supervision is dense at every frame.

\begin{table}[!t]
\centering
\caption{Comparison with a PHOC attribute-embedding query-by-string baseline on the test set. The PHOC baseline additionally receives oracle word boundaries obtained by forced alignment of the ground-truth transcription, and is therefore an upper bound for word-box-supervised methods rather than a like-for-like comparison.}
\label{tab:qbs}
\begin{tabular}{llcccc}
\toprule
Level & Method & P & R & F1 & mAP \\
\midrule
\multirow{2}{*}{Line} & PHOC (oracle boxes) & \textbf{0.664} & 0.339 & 0.449 & 0.266 \\
 & Posterior (ours) & 0.570 & \textbf{0.548} & \textbf{0.558} & \textbf{0.564} \\
\midrule
\multirow{2}{*}{Page} & PHOC (oracle boxes) & 0.601 & 0.375 & 0.462 & 0.393 \\
 & Posterior (ours) & 0.601 & \textbf{0.639} & \textbf{0.619} & \textbf{0.639} \\
\bottomrule
\end{tabular}
\end{table}

\subsection{Error Analysis}\label{sec:error}
Aggregate scores hide which queries fail and why. Table~\ref{tab:breakdown} breaks performance down by query property at $\tau^{*}$.

Query length is by far the strongest factor. Queries of 3-4 characters reach precision 0.515, while queries of 5-6 characters reach 0.861 and queries of 7 or more reach 1.000. Recall moves in the opposite direction but far more weakly, from 0.574 to 0.444. Since the median word in the dataset is only 3 characters long (Table~\ref{tab:dist}), short queries dominate the query set and drag the aggregate precision down. The mechanism is straightforward: a three-character pattern has many near-matches inside a 36-character line, and the free-start, free-end alignment is allowed to find any of them. A practical implication is that the reported aggregate understates the usefulness of the system for the queries a historian actually issues, which are typically names and technical terms of five characters or more.

The proportion of dotted letters in the query has almost no effect: F1 moves from 0.578 to 0.538 as the dotted share goes from below 25\% to above 50\%. This is worth stating because it runs against the intuition motivating the method. Posterior scoring was introduced precisely to survive dot confusions, and the flat curve is evidence that it does: the confusions are absorbed by the score rather than converted into misses. The taxonomy below supports the same reading.

\begin{table}[!t]
\centering
\caption{Word spotting performance by query property at $\tau^{*}=-0.80$.}
\label{tab:breakdown}
\begin{tabular}{lcccc}
\toprule
Query group & Queries & Precision & Recall & F1 \\
\midrule
Length 3-4 & 223 & 0.515 & 0.574 & 0.543 \\
Length 5-6 & 144 & 0.861 & 0.524 & 0.651 \\
Length 7+ & 17 & 1.000 & 0.444 & 0.615 \\
\midrule
Dotted $\leq$ 25\% & 118 & 0.580 & 0.577 & 0.578 \\
Dotted 25-50\% & 175 & 0.555 & 0.567 & 0.561 \\
Dotted $>$ 50\% & 91 & 0.540 & 0.535 & 0.538 \\
\bottomrule
\end{tabular}
\end{table}

Table~\ref{tab:fp} classifies the 635 false positives at $\tau^{*}$. A little over half (53.4\%) are cases where the query is a genuine substring of a longer word present in the line. This quantifies the limitation noted qualitatively in earlier drafts of this work: because the alignment may start and end at any frame, nothing prevents it from matching a prefix or infix. In Persian this is aggravated by clitics and by the prefixes \emph{be-}, \emph{mi-}, and \emph{na-}, which turn many short content words into substrings of longer inflected forms. A further 45.0\% are unrelated words, and only 1.6\% are words sharing the query's \emph{rasm} (skeleton) but differing in dots. The dot-confusion failure that motivates the method is therefore almost absent from the error budget, while word-boundary ambiguity accounts for the majority of it. This points to a concrete and cheap improvement: requiring a blank or space-like frame at the two ends of the match should remove a large part of the false positives without affecting recall for whole-word queries.

\begin{table}[!t]
\centering
\caption{Taxonomy of the 635 false positives of posterior search at $\tau^{*}=-0.80$.}
\label{tab:fp}
\begin{tabular}{lcc}
\toprule
Cause & Count & Share (\%) \\
\midrule
Substring of a longer word & 339 & 53.4 \\
Unrelated word & 286 & 45.0 \\
Same \emph{rasm}, different dots & 10 & 1.6 \\
\bottomrule
\end{tabular}
\end{table}

\begin{figure*}[!t]
\centering
\includegraphics[width=0.95\textwidth]{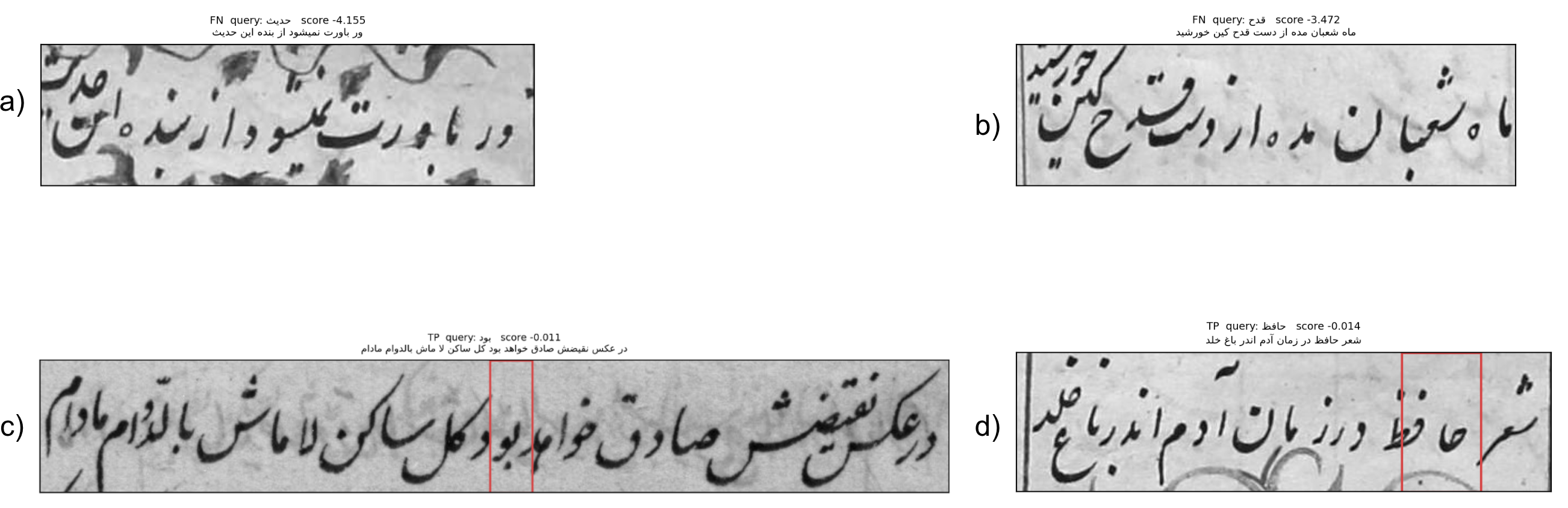}
\caption{Qualitative examples of posterior search. (a), (b): false negatives. In (a) the query \emph{hadith} is lost in a line with heavy ink blots; in (b) the short query \emph{qadah} sits in a densely joined \emph{shekasteh} hemistich where the recognizer spreads probability over several candidates. (c), (d): true positives with the word location recovered from the CTC frame alignment shown in red. No word-level annotation was used at any stage; the boxes come from the alignment alone.}
\label{fig:qualitative}
\end{figure*}

Fig.~\ref{fig:qualitative} shows four representative cases. The two false negatives, (a) and (b), are both from heavily degraded or tightly written \emph{shekasteh} lines. In (a) the query \emph{hadith} occurs at the left edge of a line with large ink blots, and the posterior is dominated by the blank symbol across those frames, so no alignment accumulates a competitive score ($-4.155$). In (b) the query \emph{qadah} is a short three-letter word inside a densely joined hemistich; the letters are present but the model distributes probability across several alternatives at every frame, leaving the per-character score at $-3.472$. Both illustrate the same failure mode: the method degrades when the recognizer is uncertain about the entire region, not when it is uncertain about one character.

The two true positives, (c) and (d), score near zero ($-0.011$ and $-0.014$), meaning the query characters were essentially the top choice at every aligned frame. The red boxes are the word locations recovered from the frame alignment, with no word-level supervision anywhere in the pipeline. In both cases the box encloses the correct word with a small horizontal margin, which is the accuracy a search interface needs in order to highlight a hit on the page.

\subsection{Performance by Page Condition}\label{sec:difficulty}
Table~\ref{tab:difficulty} reports word spotting performance on the test lines grouped by the page-level attributes of Section~\ref{sec:dataset}.

Illumination has a clear and monotone effect. Lines on pages without \emph{tazhib} reach F1 $=0.652$, lines on pages with an illuminated border drop to $0.481$, and lines on pages with both a border and a gilded cloud background drop further to $0.383$, a loss of $0.269$ F1 from the first group to the last. The cloud backgrounds are the more damaging of the two, since the gilding sits directly behind the text and reduces stroke contrast, whereas a border only affects the page margin. This is the single strongest conditioning factor we measured and it suggests that background suppression, rather than better sequence modelling, is where the next gain on this material lies.

The other two attributes behave counter-intuitively and we report them as such. Lines labelled \emph{hard} score higher (F1 $=0.576$) than lines labelled \emph{easy} ($0.507$), and lines with light degradation score higher ($0.621$) than lines with none ($0.475$). Neither ordering should be read as evidence that degradation helps. Both are consequences of how the labels are distributed: the \emph{hard} class contains 584 of the 733 test lines while \emph{easy} contains only 59, so the two are not comparable samples, and the label was assigned by reading difficulty for a human rather than by any property the model is sensitive to. The same confound affects degradation, whose categories correlate with the source books and therefore with script style and illumination rather than varying independently. We keep these rows in the table for completeness and to make the confound visible, but the only conditioning conclusion we draw from this experiment is the one about illumination. Building a difficulty axis that is both balanced and independent of the other attributes would require either a larger test set or an automatically computed score, for example from blur, contrast, and baseline curvature; we leave this to future work on the dataset.

\begin{table}[!t]
\centering
\caption{Word spotting performance on the test set (733 lines) grouped by page-level attribute, at $\tau^{*}=-0.80$. The \emph{difficulty} and \emph{degradation} groups are strongly imbalanced and confounded with script style; see the text.}
\label{tab:difficulty}
\begin{tabular}{llcccc}
\toprule
Attribute & Value & Lines & P & R & F1 \\
\midrule
\multirow{3}{*}{Tazhib} & none & 197 & \textbf{0.626} & \textbf{0.680} & \textbf{0.652} \\
 & border & 446 & 0.522 & 0.446 & 0.481 \\
 & both & 90 & 0.412 & 0.357 & 0.383 \\
\midrule
\multirow{3}{*}{Difficulty} & easy & 59 & 0.518 & 0.496 & 0.507 \\
 & moderate & 90 & 0.412 & 0.357 & 0.383 \\
 & hard & 584 & 0.585 & 0.567 & 0.576 \\
\midrule
\multirow{3}{*}{Degradation} & none & 456 & 0.512 & 0.443 & 0.475 \\
 & light & 257 & 0.607 & 0.635 & 0.621 \\
 & moderate & 20 & 0.632 & 0.590 & 0.610 \\
\bottomrule
\end{tabular}
\end{table}

\subsection{Discussion}
Four points stand out from the results.

First, fine-tuning on even a modest amount of historical Persian data is necessary. A model trained on general Arabic-script text is not enough for historical Persian manuscripts, both for finding lines and for reading them.

Second, keeping the full posterior matrix is a simple change that gives a clear gain in recall and F1 without any extra annotation or retraining, and it outperforms a PHOC attribute embedding even when that baseline is handed oracle word boundaries.

Third, the error budget is not where the motivation predicted. Dot confusions, which the method was designed to survive, account for 1.6\% of false positives, while word-boundary ambiguity accounts for 53.4\%. The method solves the problem it was built for, and the remaining precision loss is a different problem with a different and likely cheaper fix.

Fourth, illumination is the dominant page-level factor, costing 0.269 F1 between clean pages and pages with gilded cloud backgrounds.

The approach also has limitations. The word position is estimated from frames, so its horizontal boundaries are approximate, and we could not measure word-level localization accuracy directly because the dataset has no word boxes; the boxes in Fig.~\ref{fig:qualitative} are therefore illustrative rather than quantified. The precision of posterior search is limited by substring matches, which could be addressed by requiring blank frames at the match boundaries or by a lexicon or language model. Our test set of 44 pages is small enough that conditioned results on minority categories are unreliable. Finally, the subjective difficulty label did not prove to be a useful conditioning variable and should be replaced by an automatically computed one in future releases.

\section{Conclusion}\label{sec:conclusion}
We introduced a new dataset of historical Persian handwriting with 223 pages, 3,678 lines, and 37,631 words, collected from diverse books of poetry and prose and annotated at the region, line, and text level, together with page-level attributes and a distributional analysis of its content. We also proposed a word spotting baseline that is trained only with line-level labels. The method detects lines with a fine-tuned BLLA model, reads them with a fine-tuned CRNN, and searches the CTC posterior matrix instead of the decoded text. This makes the search tolerant to confusions between similar letters and gives the word position within the line and the page. On the test set, the fine-tuned line detector reaches an F1 of 0.892, and posterior search improves word spotting F1 from 0.487 to 0.558 over exact matching at a threshold selected on held-out validation data, ahead of a PHOC attribute-embedding baseline with oracle word boundaries. An error analysis shows that the dominant remaining failure is substring matching rather than character confusion, and that illuminated backgrounds are the strongest page-level factor. We hope the dataset and baseline will support further work on searching historical Persian manuscripts, for example with transformer-based recognizers \cite{Chan2024HATFormer,Sharma2026TrOCR}, learned word embeddings \cite{Krishnan2023HWNetv3}, and re-ranking methods \cite{Papazis2025Rerank}.

\section*{Data License}
The data will be distributed under the Creative Commons Attribution-NonCommercial 4.0 International (CC BY-NC 4.0) license for research purposes. The page images are derived from digitized historical manuscripts, and any use of them should also respect the terms of the holding institutions.


\begin{thebibliography}{18}

\bibitem{Sadri2016}
J.~Sadri, M.~R. Yeganehzad, and J.~Saghi, ``A novel comprehensive database for offline Persian handwriting recognition,'' \emph{Pattern Recognit.}, vol.~60, pp.~378-393, 2016.

\bibitem{Jafarzadeh2024Khayyam}
P.~Jafarzadeh, P.~Choobdar, and V.~Mohammadi Safarzadeh, ``Khayyam offline Persian handwriting dataset,'' arXiv preprint arXiv:2406.01025, 2024.

\bibitem{Khosravi2007Hoda}
H.~Khosravi and E.~Kabir, ``Introducing a very large dataset of handwritten Farsi digits and a study on their varieties,'' \emph{Pattern Recognit. Lett.}, vol.~28, no.~10, pp.~1133-1141, 2007.

\bibitem{Ziaratban2009FHT}
M.~Ziaratban, K.~Faez, and F.~Bagheri, ``FHT: An unconstraint Farsi handwritten text database,'' in \emph{Proc. 10th Int. Conf. Document Anal. Recognit. (ICDAR)}, 2009, pp.~281-285.

\bibitem{Marti2002IAM}
U.-V. Marti and H.~Bunke, ``The IAM-database: An English sentence database for offline handwriting recognition,'' \emph{Int. J. Document Anal. Recognit.}, vol.~5, pp.~39-46, 2002.

\bibitem{Kassis2017VMLHD}
M.~Kassis, A.~Abdalhaleem, A.~Droby, R.~Alaasam, and J.~El-Sana, ``VML-HD: The historical Arabic documents dataset for recognition systems,'' in \emph{Proc. 1st Int. Workshop Arabic Script Anal. Recognit. (ASAR)}, 2017, pp.~11-14.

\bibitem{Basharat2026UrduKatib}
R.~Basharat and M.~U. Ali, ``Urdu Katib handwritten dataset: A historical document dataset for offline Urdu handwritten text recognition with CRNN-based baseline evaluation,'' arXiv preprint arXiv:2606.19139, 2026.

\bibitem{Allen2026Makhzan}
J.~P. Allen \emph{et al.}, ``OpenITI MAKHZAN: An open annotated dataset of Arabic, Persian, Ottoman Turkish, and Urdu print and manuscript data,'' \emph{J. Open Humanities Data}, vol.~12, art.~69, pp.~1-12, 2026, doi: 10.5334/johd.465.

\bibitem{Jampour2026MPHD}
M.~Jampour, S.~Farridnejad, A.~KarimiSardar, K.~Champour, A.~Aghaee Meybodi, F.~Hematzadeh, and L.~Paul, ``MPHD: Multi-purpose Persian handwriting dataset for text recognition, line segmentation, writer identification, and handwritten analysis,'' preprint, 2026, doi: 10.17632/2fcw6cd9gd.1.

\bibitem{Kiessling2020BLLA}
B.~Kiessling, ``A modular region and text line layout analysis system,'' in \emph{Proc. 17th Int. Conf. Frontiers Handwriting Recognit. (ICFHR)}, 2020, pp.~313-318.

\bibitem{Chan2024HATFormer}
A.~Chan, A.~Mijar, M.~Saeed, C.-W. Wong, and A.~Khater, ``HATFormer: Historic handwritten Arabic text recognition with transformers,'' arXiv preprint arXiv:2410.02179, 2024.

\bibitem{Wang2017GRCNN}
J.~Wang and X.~Hu, ``Gated recurrent convolution neural network for OCR,'' in \emph{Adv. Neural Inf. Process. Syst. (NIPS)}, 2017, pp.~335-344.

\bibitem{Almazan2014Attributes}
J.~Almaz\'{a}n, A.~Gordo, A.~Forn\'{e}s, and E.~Valveny, ``Word spotting and recognition with embedded attributes,'' \emph{IEEE Trans. Pattern Anal. Mach. Intell.}, vol.~36, no.~12, pp.~2552-2566, 2014.

\bibitem{Sudholt2016PHOCNet}
S.~Sudholt and G.~A. Fink, ``PHOCNet: A deep convolutional neural network for word spotting in handwritten documents,'' in \emph{Proc. 15th Int. Conf. Frontiers Handwriting Recognit. (ICFHR)}, 2016, pp.~277-282.

\bibitem{Wilkinson2017CtrlF}
T.~Wilkinson, J.~Lindstr\"{o}m, and A.~Brun, ``Neural Ctrl-F: Segmentation-free query-by-string word spotting in handwritten manuscript collections,'' in \emph{Proc. IEEE Int. Conf. Comput. Vis. (ICCV)}, 2017, pp.~4433-4442.

\bibitem{Papazis2025Rerank}
S.~Papazis, A.~P. Giotis, and C.~Nikou, ``Enhancing keyword spotting via NLP-based re-ranking: Leveraging semantic relevance feedback in the handwritten domain,'' \emph{Electronics}, vol.~14, no.~14, art.~2900, 2025.

\bibitem{Krishnan2023HWNetv3}
P.~Krishnan, K.~Dutta, and C.~V. Jawahar, ``HWNet v3: A joint embedding framework for recognition and retrieval of handwritten text,'' \emph{Int. J. Document Anal. Recognit.}, vol.~26, pp.~401-417, 2023.

\bibitem{Sharma2026TrOCR}
S.~Sharma, M.~Flammini, and F.~Simonetta, ``TrOCR for medieval HTR: A systematic ablation study with cross-dataset validation,'' arXiv preprint arXiv:2606.24302, 2026.

\end{thebibliography}
\end{document}